# FUZZY CLUSTERING BASED SEGMENTATION OF VERTEBRAE IN T1-WEIGHTED SPINAL MR IMAGES

Jiyo.S.Athertya and G.Saravana Kumar

Department of Engineering Design, IIT-Madras, Chennai, India

*ABSTRACT*

*Image segmentation in the medical domain is a challenging field owing to poor resolution and limited contrast. The predominantly used conventional segmentation techniques and the thresholding methods suffer from limitations because of heavy dependence on user interactions. Uncertainties prevalent in an image cannot be captured by these techniques. The performance further deteriorates when the images are corrupted by noise, outliers and other artifacts. The objective of this paper is to develop an effective robust fuzzy C- means clustering for segmenting vertebral body from magnetic resonance image owing to its unsupervised form of learning. The motivation for this work is detection of spine geometry and proper localisation and labelling will enhance the diagnostic output of a physician. The method is compared with Otsu thresholding and K-means clustering to illustrate the robustness.The reference standard for validation was the annotated images from the radiologist, and the Dice coefficient and Hausdorff distance measures were used to evaluate the segmentation.*

## 1.INTRODUCTION

Image segmentation is a fundamental building block in an image analysis tool kit. Segmentation of medical images is in itself an arduous process where the images are prone to be affected by noise and artifacts. Automatic segmentation of medical images is a difficult task as medical images are complex in nature and rarely posses simple linear feature characteristic. Further, the output of segmentation algorithm is affected due to partial volume effect, intensity inhomogeneity in case of magnetic resonance (MR) images.

Spine is the most complex load bearing structure in our entire human body.It is made up of 26 irregular bones connected in such a way that flexible curvedstructure results. The vertebral column is about 70cm long in an average adult and has5 major divisions. Seven vertebrae found in the neck region, constitutethe cervical part, the next 12 are the thoracic vertebrae and 5 supporting the lower back are the lumbar vertebrae. Inferior to these, is the sacrum that articulates with the hipbones of pelvis. The tiny coccyx terminates the entire column. Intervertebral discacts as a shock absorber and allow the spine to extend. These are thickest inthe lumbar and cervical regions, enhancing the flexibility in these regions. Its degenerationis relatively a common phenomena with aging due to wear and tear and is the major cause for back pain[1]. Herniated disc, spinalstenosis and degenerative discs are a few of the types, to mention. These can be imaged andstudied from MRI scans and this modality of imaging is prescribed most commonly for patients with excruciatingback pain. MR imaging of spine is formallyidentified with IR (Inversion Recovery), T1and T2 weighted images. While water content appears bright in T2(in medical lingo, its hyper intense which is clearly seen in the spinal canal), the same appears dark (hypo intense) in T1 images.MR can detectearly signs of bone marrow degeneration with high spatial resolution where fat and water protons are found in abundance. These changes named as Modic changes can be diagnosed using MR imaging [1]. Degenerated L5 vertebrae and the associated intensity changes that are prevalent particularly in end plates are shown in fig 1. While degenerative changes are a biological phenomena occurring in spinal structure that are imaged using radiological equipment, certain irrelevant processes are also captured. These constitute the artefacts caused due to intensity inhomogeneity as shown in fig 2. The segmentation process is highly affected by these complexities present in MR images. Accurate diagnosis remains a challenge without manual intervention in segmenting the vertebral features. Robust automatic segmentation of vertebrae from MR images would be a pre requisite for diagnosis using computer methods. The current work deals with segmentation of vertebrae from MR image using fuzzy c-means (FCM) clustering for identification and labelling of individual vertebral structures. The segmented output can be refined further and used for classification of degenerative state as well as to diagnose deformities.

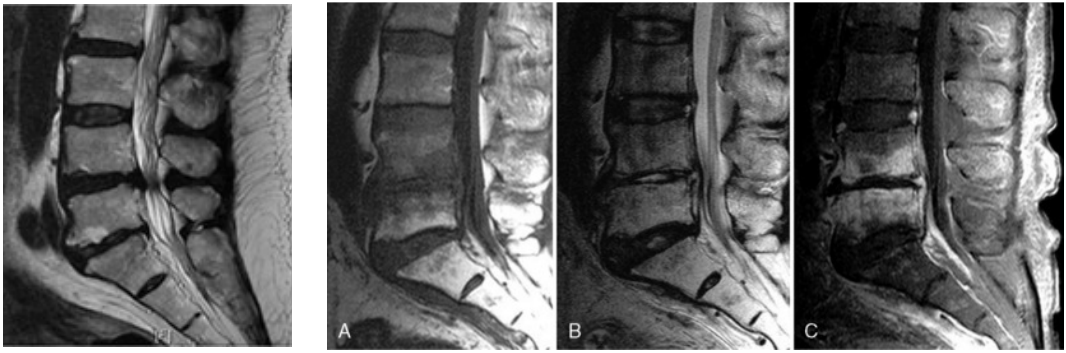

Figure 1. Degenerated L5 vertebra in MR sagittal plane

Figure 2. Intensity inhomogeneity captured in lumbar vertebrae

## 2. LITERATURE

The commonly used segmentation methods are global thresholding, multilevel thresholding and supervised clustering techniques. In intensity thresholding, the level determined from the grey-level histogram of the image. The distribution of intensities in medical images, especially in MRI images is random, and hence global thresholding methods fail due to lack of determining optimal threshold. In addition, intensity thresholding methods have disadvantage of spatial uncertainty as the pixel location information is ignored [2]. An edge detection scheme can be used for identifying contour boundaries of the region of interest (ROI). The guarantee of these lines being contiguous is very sleek. Also, these methods usually require computationally expensive post-processing to obtain hole free representation of the objects.

A graph based approach for MRI vertebral body segmentation is proposed in [3].Here anNyström approximation is used for reducing the processing speed of normalized cut (NCut) method. These methods maximise the similarity within a subset and dissimilarity between two subsets in an image. Proposed by Malik and Shi, the problem reduces to solving the generalised Eigen vector that involves a diagonal matrix and a weighted matrix that is formed using the feature strength of the image. The computationally intensive calculation of the Eigen vectors is simplified using Nyström approximation that calculates the diagonal matrix using few random samples and then extends the solution to the complete set. Pre-processing is done using anisotropic diffusion that provides a coil corrected output.

A semiautomatic classification of spine disorders namely disc herniation, degeneration and spinal stenosis is presented in [4]. For the case of disc degeneration classification, the authors have captured MRI response in T2 weighted images. A decrease in mean intensity denotes the level of degeneration and is classified via Pfirmann scale. Disc segmentation is done using gradient vector flow (GVF) on the pre-processed imagesfollowed by skeletonization. From the mid-point, a vertical axis is drawn, to mark the length. Changes in length as well as intensity leads to classification of abnormalities in disc. Since 90% of herniation [5] happen in Lumbar region, the processing deals with discs located in the lumbosacral area. For disc herniation, evaluation is done on contour extracted from axial slice. The classification of herniated discs is either extrusion or protrusion(focal based or broad based).

A quasi automatic segmentation method for intervertebral disc (IVD) is given in [6]. The only intervention being, initial point selection for evolving an elliptical contour to be deformed by snakes. The final refinement of region of interest is done using FCM. A new energy term called the geometric energy is included with active contour evolving equation that models the shape of disc and restricts it to a certain limit. In this case, since IVD resemble elliptical structure, the contour evolution is curtailed when it goes beyond the mentioned shape. Classification of discs is done using Adaboost classifier while validation is performed using Dice similarity coefficient and root mean square error (RMSE) between ground truth and segmented disc.

The region growing methods extend the thresholding by integrating it with connectivity by means of an intensity similarity measure. These methods assume an initial seed position and using connected neighbourhood, expand the intensity column over surrounding regions. However, they are highly sensitive to initial seeds and noise. In classification-based segmentation method, FCM algorithm [7], is more effective with considerable amount of benefits. Unlike hard clustering methods, like K-means algorithm, which assign pixels exclusively to one cluster, the FCM algorithm allows pixels to have dependence with multiple clusters with varying degree of memberships and thus more reasonable in real applications. Using intuitionistic fuzzy clustering (IFC), where apart from membership functions (MF), non membership values are also defined, [8]have segmented MR images of brain. The heuristic based segmentation also considers the hesitation degree for each pixel. A similar study on generic grey scale images is put forth in [9] where the IFC combines several MF's to address the uncertainty in choosing the best MF.

Detection of Modic changes (MC) can significantly contribute to the lesion detection on the bone marrow. It can also be associated with disc herniation [1]. It has been shown that the higher the severity of Modic changes, greater is level of herniation. Automating the process of classification and detection can in turn lessen the burden of manual demarcation of MC levels for a radiologist. It can as well provide an accurate percentage of degeneration based on intensity change in signal compared to the normal level.While a very few articles have been reported in the MR segmentation of vertebral bodies [10][11][12] because of the complexities involved in delineating VB's, our focus is on utilising the fuzzy logic for accurately isolating them. The article proposes a FCM based method to segment vertebral bodies (VB) with morphological post processing. Also the VB's are labelled which can reduce the burden of radiologist while classifying the degenerations involved.

## 3. METHODS

The proposed method is schematically depicted in fig. 3. The input image(s) have been collected from Apollo Speciality Hospitals,Chennai after going through a formal ethical clearance process. The T1 weighted images, served as the initial dataset for the proposed algorithm. Initially the image is smoothed using the edge preserving anisotropic diffusion filter. This pre-processing serves the dual purpose of removing inhomogeneity and as an enhancer as well.

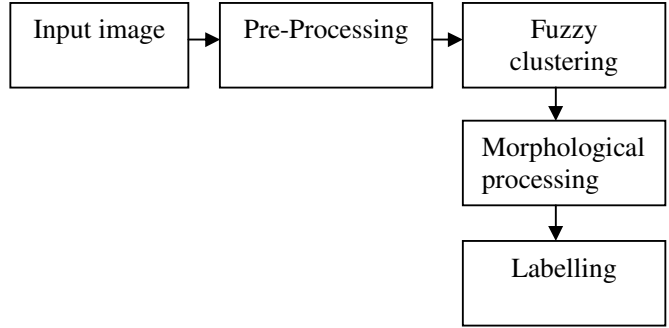

Figure 3. Schematic of the proposed segmentation method

### 3.1. Fuzzy C-Means Clustering

The FCM[2]has been broadly used invarious pattern and image processing studies [10]–[12]. According to FCM algorithm, the clustering of a dataset can be obtained by minimizing an objective function *J* for a known number of clusters:

$$J = \sum_{i=1}^{N}\sum_{j=1}^{M} u_{ij}^{k} \|x_i - v_j\|^2, \ 1 \leq k < \infty \tag{1}$$

where ;

$k$ is any real number known as the weighting factor, $u_{ij}$is degree of membership of $x_i$ in the cluster *j*, $x_i$is the$i^{th}$of *p*-dimensional measured intensity data, $v_j$ is the *p*-dimensional center of the $j^{th}$cluster, $\|*\|$ is any norm expressing the similarity between measured intensity data and center, and *N* represents number of pixels while *M* represents the number of cluster centers. Fuzzy clustering is performed through an iterative optimisation of objective function shown in Eq. (1) with update of membership function $u_{ij}$and cluster centers$v_j$ as given in Eq. (2). The algorithm is terminated when max$_{ij}$\{$u_{ij}$ at *t+1* - $u_{ij}$at *t*\} $\leq \epsilon$ which is between 0 and 1.

$$u_{ij} = \frac{1}{\sum_{l=1}^{M} \left(\left\|\frac{x_i - v_j}{x_i - v_l}\right\|\right)^{\frac{2}{(k-1)}}}$$

$$v_j = \frac{\sum_{i=1}^{N} u_{ij}^{k} x_i}{\sum_{i=1}^{N} u_{ij}^{k}} \tag{2}$$

### 3.2. Post Processing and Labelling

A series of morphological operations are executed for extracting the vertebral bodies (VB) from the clustered output. Hole filling is the preliminary step followed by an erosion to remove islands. An area metric is used to extract only Vertebrae from surrounding muscular region Shape analysis [13] reveals that the aspect ratio of VB varies between 1.5 and 2. This helps in isolating the ligaments and spinal muscles associated with the spine in the region of interest.The segmented vertebrae are labelled using the connected component entity. Each VB is identified with a group number. Starting from L5 (Lumbar), the vertebrae are labelled successively till L1. If the sacrum remains due to improper segmentation, it can be eliminated based on aspect ration or area criteria. A colour schematic is also presented for visual calibration.

### 3.3. Validation

The proposed method was validated using Dice coefficient (DC) and Hausdorff distance (HD). The reference standard for comparison was the annotated images from the radiologist. DC measures the set agreement as described in following equations, where the images constitute the two sets. The generalized HD provides a means of determining the similarity between two binary images. The two parameters used for matching the resemblance between the given images are,

- Maximum distance of separation between points, yet that can still be considered close.
- The fraction that determines how much one point set is far apart from the other.

$$D(A, B) = \frac{2|A \cap B|}{|A| + |B|} \text{ (Dice Coefficient)}$$

$$D(A, B) = \underbrace{Max}_{a \in A} \{\underbrace{Min}_{b \in B} \{d(a, b)\}\} \text{ (Hausdorff Distance)}$$

where, *a, b* are points from the images *A,B* respectively.

## 4. RESULTS AND DISCUSSION

The method is tested on sagittal cross-section of T1-weighted MR images of spine.The goal is to segment the vertebral bodies from the muscular background and label them accordingly.In total 16 patient data (MR images) were used for this study. The patients complained of mild lower back pain and are in the age group between 45-60. The population included 8 female and 8 male. The results from FCM were compared with the segmentation using Otsu thresholding and K-means clustering is presented in the subsequent sections.

### 4.1. Illustration of Methodology

One of the patients MR sagittal slice of spine considered for the current study is shown in fig 4(a) and the proposed methodology of FCM for segmentation and labelling of vertebrae is illustrated in fig 4(b-d). After the pre-processing stage, the enhanced input is clustered using the FCM technique followed by a set of morphological operations (the results are shown in fig. 4(b)).Automatic labeling of vertebrae is usually performed to reduce the manual effort put in by the radiologist. It can be seen from fig 4(c) and (d), the labeled vertebrae and its color scheme can help in better diagnosis. The intermediate steps involve various morphological operations that are depicted in fig 5. It can be seen that, the fuzzy clustering provides a closer disjoint VB's owing to

International Journal of Fuzzy Logic Systems (IJFLS) Vol.6, No.2,April 2016

which one can erode the muscular region and thus arrive at delineating the same. Also shown in fig 5. are the output obtained using Otsu thresholding and K means clustering with the intermediate steps. An image overlay of the input and segmented output using FCM for all the 16 cases considered for the study is presented in fig 6.

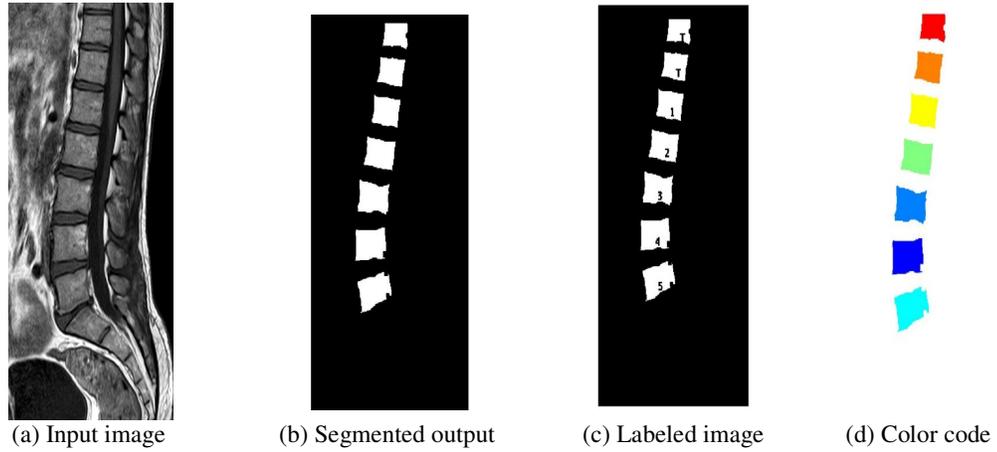

  (a) Input image      (b) Segmented output     (c) Labeled image     (d) Color code

Figure 4. MR segmentation and labeling.

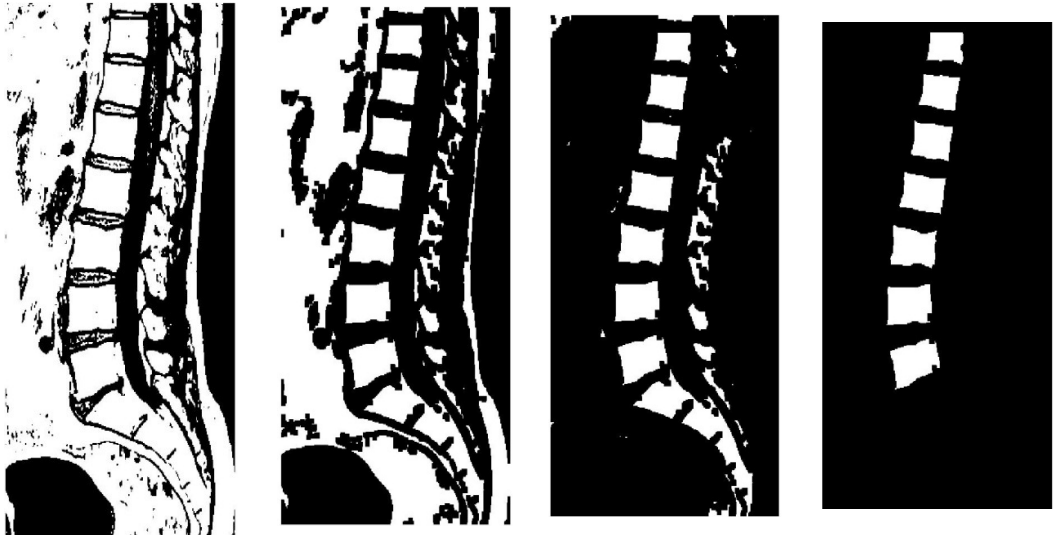

Fuzzy c-means clustering

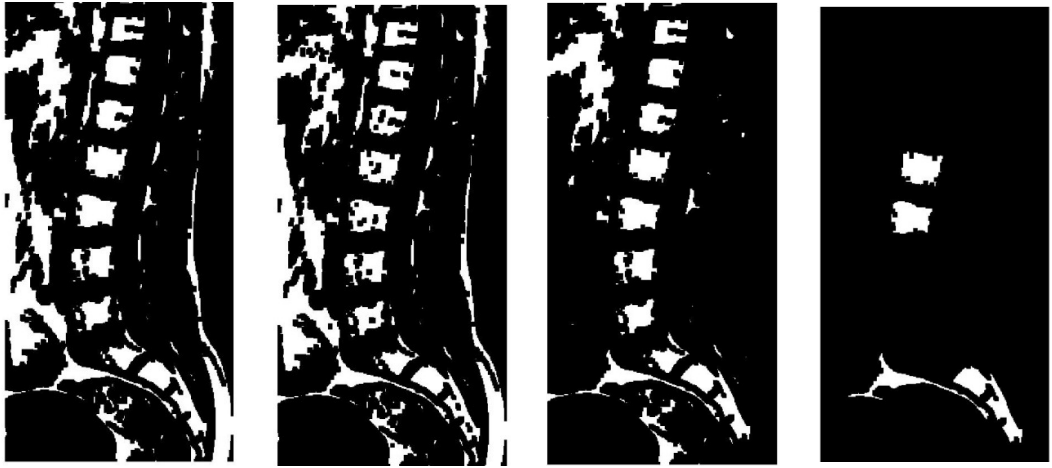

Otsu's thresholding

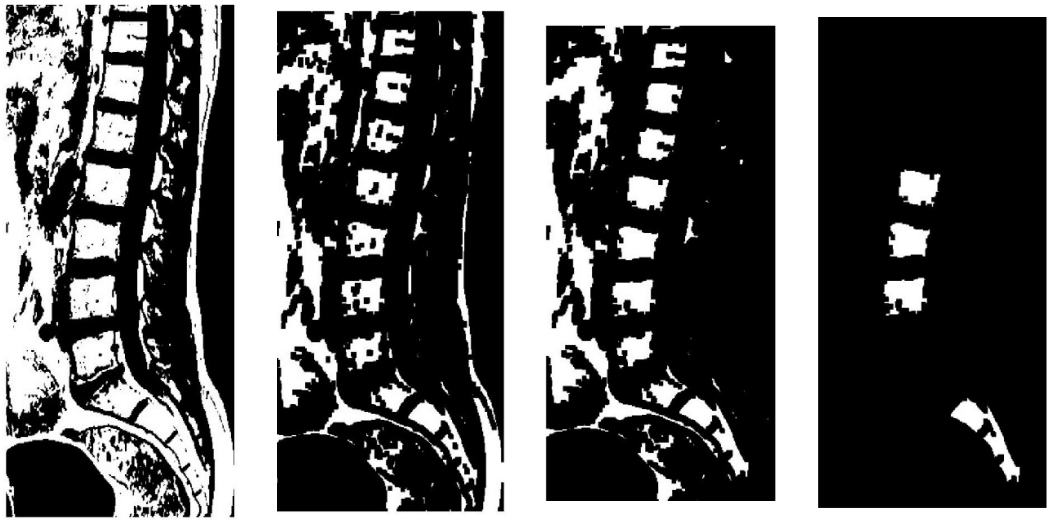

K-means clustering

(a) Binary label     (b) Erosion     (c) Filtering using area criteria     (d) Aspect ratio based elimination

Figure 5. Comparative illustration of segmentation using fuzzy c-means, Otsu and K-means.

29

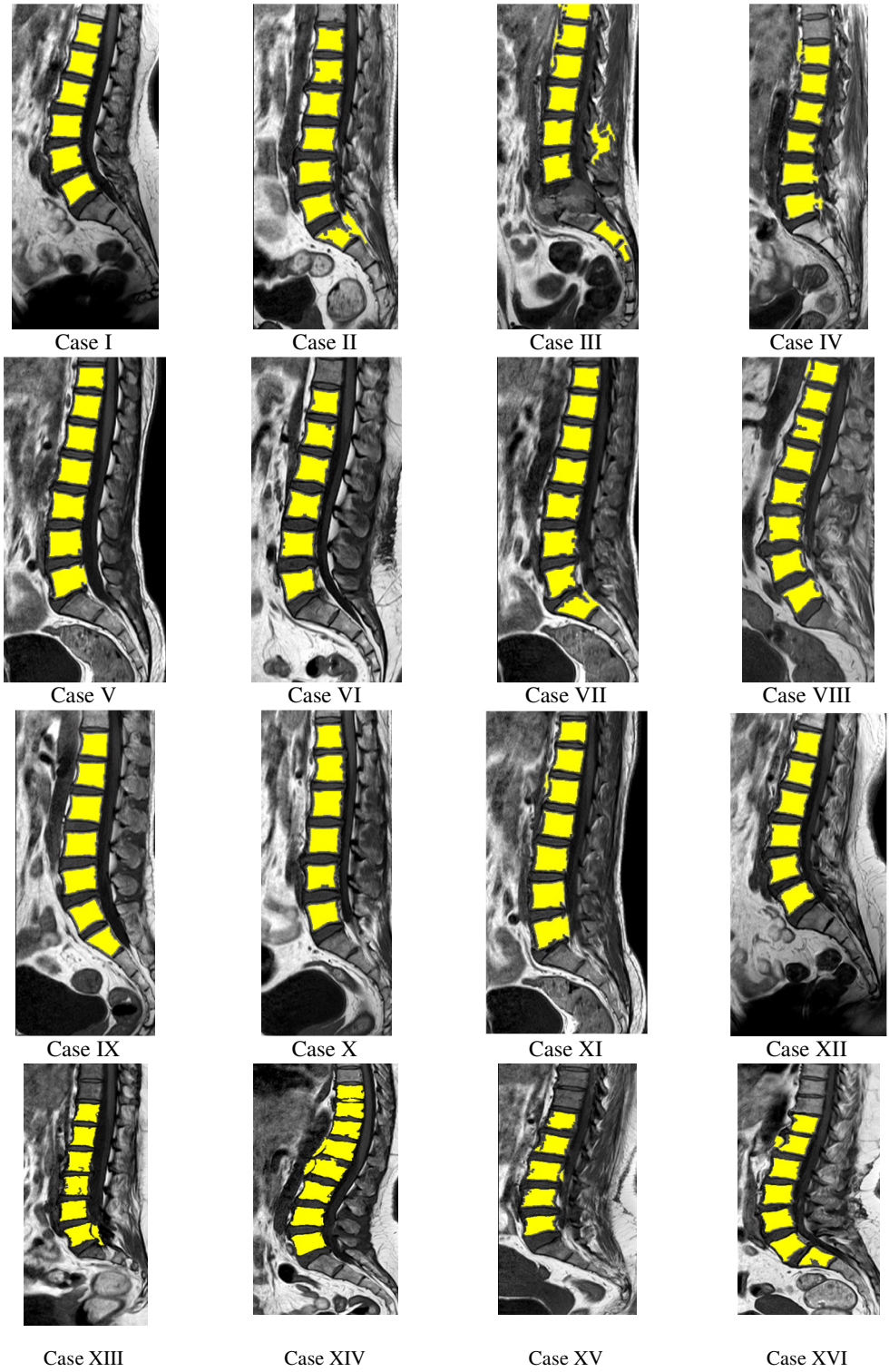

Figure 6. Overlay of segmented image with input for various case studies

## 4.2. Comparative Analysis

A comparative studyon performances of Otsu thresholding, k-means clustering and the FCM was conducted based on DC and HD metric as compared to the gold standard segmentation by radiologist. While a DC metric should be as close to unity, HD on the contrary, has to be low for achieving accurate segmentation results. Statistical analysis of these metric for the results obtained for the 16 cases considered for the study was performed. A descriptive statistical analysis was carried and the results tabulated in table I and II. It can be observed that the Fuzzy method provides better DC value (closer to 1) and HD value (closer to 0) than compared to the rest thus affirming the robustness in segmentation. The result of the study is illustrated in the form of box plot in figures 7 and 8. It is inferred from the plot that fuzzy clustering outperforms the other two methods chosen for comparison. FCM also shows lower variance and thus indicating robustness. Otsu thresholding has a relatively large variance in DC metric. In the case of HD metric, K-means has a larger variance.

| Table I . Descriptive statistics for DC Validation metric | | | | |
|---|---|---|---|---|
| | N | Mean | SD | SEM |
| Otsu thresholding | 16 | 0.31674 | 0.14542 | 0.03635 |
| K means clustering | 16 | 0.78621 | 0.05227 | 0.01307 |
| Fuzzy c means clustering | 16 | 0.8672 | 0.0407 | 0.01017 |

| Table II . Descriptive statistics for HD Validation metric | | | | |
|---|---|---|---|---|
| | N | Mean | SD | SEM |
| Otsu thresholding | 16 | 13.76807 | 2.80003 | 0.70001 |
| K means clustering | 16 | 8.5913 | 2.56782 | 0.64195 |
| Fuzzy c means clustering | 16 | 5.40424 | 1.12019 | 0.28005 |

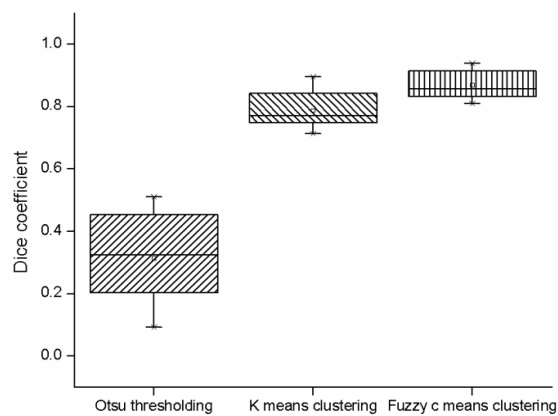

Figure 7. Comparative analysis of methods using Dice coefficient

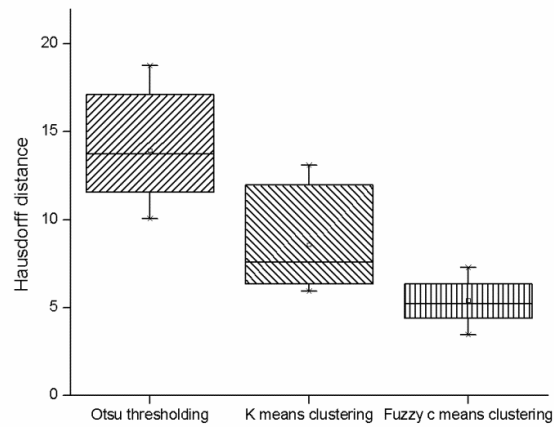

Figure 8. Comparative analysis of methods using Hausdorff distance

### 4.3. Time Complexity

The computational time taken for the proposed algorithm is compared against time expended by Otsu thresoldingand K-means clustering and tabulated in table III. It can be seen that mean time for fuzzy clustering is almost 5 times the time taken by Otsu thresholding and K-means clustering. However, since the operation is executed in seconds, it is deduced that additional computational burden might not play a significant role and is justified due to the higher quality of results obtained using FCM. The result of the analysis are also shown as box plot in figure 9.

| Table III . Descriptive statistics for computational time | | | | |
|---|---|---|---|---|
| | N | Mean | SD | SEM |
| Otsu thresholding | 16 | 0.00235 | 8.26735E-4 | 2.06684E-4 |
| K means clustering | 16 | 1.27684 | 0.35717 | 0.08929 |
| Fuzzy c means clustering | 16 | 5.61202 | 0.80793 | 0.20198 |

### 4.4. Failure Case

The method was tested on several images and in some images the segmentation failed to provide quality results. The transverse and spinous processes are a part of vertebral bodies. Thus, when they start emerging, with disruption in intensity as well as structure, the fuzzy clustering method fails to adapt to the complex topology. Apart from this, the presence of anterior and posterior ligaments also significantly affects the results of the segmentation. Fig. 10. shows the results of segmentation of one such case where the ROI has not been delineated clearly. It should be noted that the other competing methods considered for the study namely, Otsu thresholding as well as K-means clustering were also not successful in proving the segmentation in these cases.

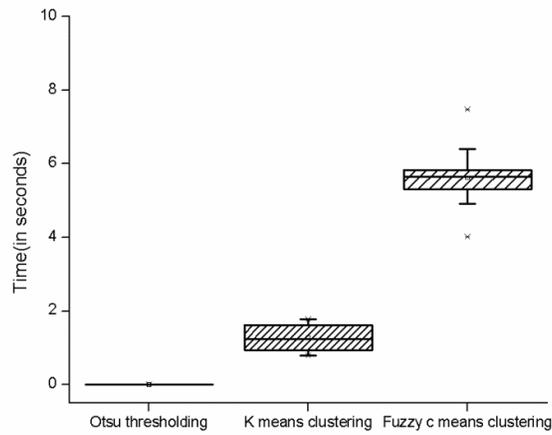

Figure 9. Box plot featuring the computational time elapsed (in seconds)

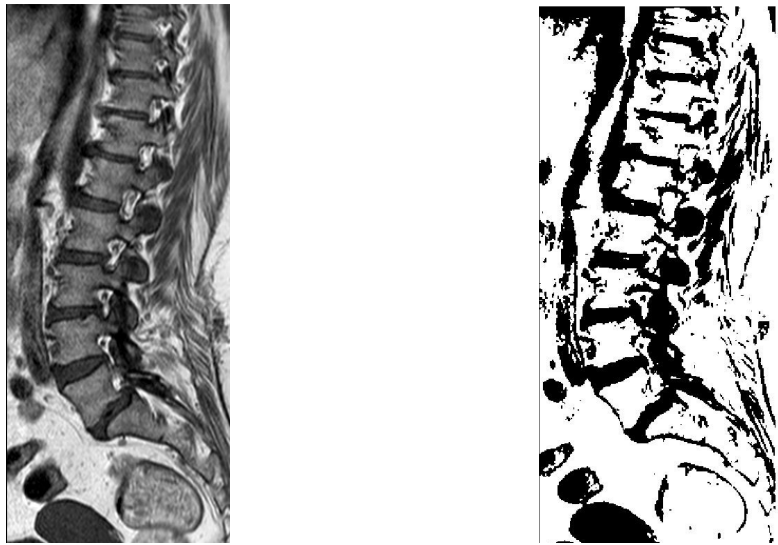

Figure 10. A failure case of proposed segmentation

## 5. CONCLUSIONS

In this paper, a FCM algorithm followed by morphological operations and labelling has been presented for segmentation of vertebra from spine MR images. It is compared with the simple K-means clustering and Otsu thresholding scheme. The study included 16 patient dataset constituting 8 female and 8 male T1-weighted MR images of spine for comparison and validation. Upon validation, it is observed that the FCM method gives improved segmentation results as compared to the counterparts. Time complexity of the methods is also presented. As a part of

33

future work, we would like improve the method by incorporating intuitionistic fuzzy clustering scheme and also extract features from the segmented VB for classifying various deformity.

## ACKNOWLEDGEMENTS

The first author would like to thank the Department of Science and Technology [DST], India, for supporting the research through INSPIRE fellowship